\documentclass[10pt,twocolumn,letterpaper]{article}

\usepackage{cvpr} 

\usepackage[dvipsnames]{xcolor}
\usepackage{multirow}

\definecolor{cvprblue}{rgb}{0.21,0.49,0.74}
\usepackage[pagebackref,breaklinks,colorlinks,citecolor=cvprblue]{hyperref}

\title{Neural Modes: Self-supervised Learning of Nonlinear Modal Subspaces}

\author{Jiahong Wang, Yinwei Du, Stelian Coros, Bernhard Thomaszewski\\
ETH Z\"{u}rich\\
{\tt\small \{wangjiah, duyin, scoros, bthomasz\}@ethz.ch}
}
\usepackage{xspace}
\makeatletter
\DeclareRobustCommand\onedot{\futurelet\@let@token\@onedot}
\def\@onedot{\ifx\@let@token.\else.\null\fi\xspace}

\makeatother

\newcommand{\bE}{\mathbf{E}}

\newcommand{\bH}{\mathbf{H}}

\newcommand{\bM}{\mathbf{M}}

\newcommand{\bX}{\mathbf{X}}

\newcommand{\be}{\mathbf{e}}

\newcommand{\bn}{\mathbf{n}}

\newcommand{\bu}{\mathbf{u}}

\newcommand{\bx}{\mathbf{x}}

\newcommand{\by}{\mathbf{y}}
\newcommand{\bz}{\mathbf{z}}
\newcommand{\bzero}{\mathbf{0}}

\usepackage{amsmath}

\begin{document}

\maketitle
\begin{abstract}

We propose a self-supervised approach for learning physics-based subspaces for real-time simulation. Existing learning-based methods construct subspaces by approximating pre-defined simulation data in a purely geometric way. However, this approach tends to produce high-energy configurations, leads to entangled latent space dimensions, and generalizes poorly beyond the training set. To overcome these limitations, we propose a self-supervised approach that directly minimizes the system's mechanical energy during training. We show that our method leads to learned subspaces that reflect physical equilibrium constraints, resolve overfitting issues of previous methods, and offer interpretable latent space parameters.
\end{abstract}    

\begin{figure}[t]
\begin{minipage}{\columnwidth}
    \centering
    \includegraphics[width=\linewidth]{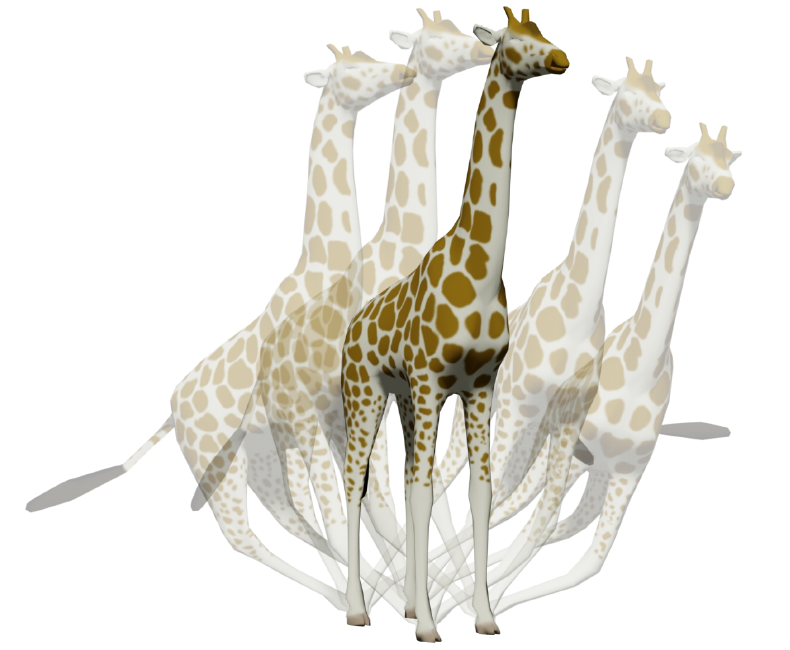}
    \caption{Visualization of our nonlinear subspace on a giraffe model discretized with tetrahedron finite elements. Overlays correspond to different frames from a selected neural mode.}
    \label{fig:teaser}
    \vspace{-0.5cm}
\end{minipage}
\end{figure}

\section{Introduction}
Physics-based simulation is an indispensable tool in science and engineering. Computer graphics has since long relied on simulation to create stunning visual effects and animations. However, while our ability to create ever more realistic simulations is steadily increasing, physics-based simulations are notoriously time-consuming, and translating techniques to the real-time regime remains a major challenge. 
One particular strategy to reduce computational burden without sacrificing visual quality is model reduction \cite{hahn2014subspace, vijayaraghavan2023data, li2023subspace}. These techniques construct low-dimensional subspaces that aim to capture the variability and expressiveness of full-space simulations. However, designing subspaces that strike a balance between visual fidelity and computational complexity is no easy task. While linear vibration modes are ideal for small deformations, they fail to capture motions involving large deformations and rotations.

Instead of approximating nonlinear simulations with a linear subspace, recent work has started to explore machine learning techniques to build nonlinear subspaces from simulation data \cite{Fulton:LSD:2018}.
While this learning-based approach to subspace simulation has shown its promise, existing methods rely on supervised training, which leads to a number of challenges and limitations. First, supervised learning requires significant amounts of pre-defined and curated training data. Second, this data is typically approximated in a purely geometric way, e.g., by minimizing the reconstruction loss of a deep autoencoder (AE) \cite{KRAMER1992313}. This geometric approach, however, ignores the fact that simulation data live on a nonlinear manifold defined through mechanical equilibrium conditions. Consequently, configurations from geometrically-learned subspaces generally exhibit much higher internal energy than the input data, which manifests as stress concentrations and spurious forces that affect the quality and performance of simulations. Another problem is the empirical observation that supervised learning often leads to overfitting and generalizes poorly to configurations outside the training set. We confirm this observation for the case of nonlinear subspace simulation through a series of experiments. Finally, the learned representation of the subspace typically lacks structure and is often strewn with singularities and discontinuities. In particular, latent space dimensions are rarely meaningful (neither from energetic, visual, nor semantic points of view), and smooth trajectories in latent space typically lead to incoherent geometry.

To address these challenges, we propose a novel self-supervised approach to learning nonlinear subspaces. Rather than building geometry-based subspaces from simulation data, our method learns physics-based subspaces by directly minimizing the energy of the underlying mechanical system. 
We draw inspiration from Nonlinear Compliant Modes, an extension of linear modes to the large deformation regime proposed by D{\"u}nser et al. \cite{Duenser22NCM}. Each nonlinear mode is parameterized by a modal coordinate defined as the projection of a given configuration onto the corresponding linear mode. For any modal coordinate value, the corresponding nonlinear mode shape is obtained by minimizing the elastic energy in the space orthogonal to the linear mode. 
To construct our neural subspaces, we extend Nonlinear Compliant Modes to nonlinear modal spaces that capture the behavior along and across a given set of modes.
Crucially, since the map between modal coordinates and nonlinear equilibrium shapes is given by a constrained minimization problem, we can train our neural network to learn the entire solution manifold by minimizing the corresponding energy across parameter space.
Once trained, our method allows for real-time exploration and gradient-based navigation of nonlinear modal subspaces. We demonstrate the potential of our approach on a range of examples spanning real-time dynamics simulation, physics-based nonlinear deformations, and keyframe animation.
We furthermore analyze the performance of our method through ablation studies and comparisons to baselines, revealing that our self-supervised approach outperforms existing methods based on supervised learning.
In summary, our work makes the following contributions:
\begin{itemize}
    \item We present the first self-supervised approach for learning nonlinear modal subspaces for simulation. Our approach eliminates the need for creating curated data collections, enables end-to-end physics-based training, and has quantified physical accuracy.
    \item We extend nonlinear compliant modes from single modes to fully coupled modal spaces for exploration.
    \item We analyze the quality of our learned subspaces and existing alternatives with respect to accuracy, smoothness, and interpretability. While existing
    data-free methods \cite{sharp2023datafree} can lead to mode collapse, we show that our formulation avoids this problem. We further compare the performance of our solution to baselines on a set of subspace simulation examples.
\end{itemize}

\section{Related work}

\textbf{Subspace Simulation.}
Model reduction or subspace simulation is an established technique with broad applications in many scientific disciplines. Arguably, the most widely used approach is linear modal analysis \cite{shabana1991theory}, initially designed for vibration problems. The basic idea of this approach is to define a low-dimensional linear basis using linear modes, i.e., eigenvectors of the energy Hessian corresponding to low-frequency deformations. Despite its efficiency for small deformations, linear eigenmodes are known to suffer from severe distortion artifacts for large deformations  \cite{10.1145/2019627.2019638}.

Various methods aimed to reduce artefacts from linear eigenmodes by augmenting the basis with carefully selected vectors. Some notable examples include modal derivatives \cite{10.1145/1073204.1073300},  higher-order Krylov-type modes \cite{10.1145/2816795.2818089}, higher-order descent directions \cite{10.1145/2019627.2019638}, modal warping that corrects for rotations \cite{10.1109/TVCG.2005.13}, and rotation strain extrapolation \cite{rotstrain}. 
Nonlinear compliant modes \cite{Duenser22NCM} are arguably the approach most closely related to our work. These modes are found by constraining the projection of the system's configuration onto the linear eigenmodes while minimizing its internal energy in the orthogonal subspace. The method is fully physics-based and has been demonstrated for various analysis tasks on real-world 3D printed examples. However, in their original formulation, nonlinear compliant modes are inherently one-dimensional and do not readily extend to multi-dimensional subspace. We extend the concept of nonlinear compliant modes to complete modal subspaces, enabling exploration and simulation.

\textbf{Nonlinear Subspaces.}
While the majority of works focus on modal analysis, alternative strategies for constructing nonlinear subspace have been explored as well. Rig-space approaches solve forward simulation and inverse kinematics tasks in the nonlinear subspaces induced by character rigs \cite{10.1145/2185520.2185568, 10.1145/2485895.2485918, rig3, PhysIK}. However, these methods are tailored to articulated objects and rely on artist-made rigs.

Example-based approaches \cite{example1, example2, 10.1145/1531326.1531359, pose-space} build subspace based on a few user-selected example states. Increasing the number of examples to full simulation sequences gives rise to data-driven methods. A widely used approach in this context is to construct linear subspaces from simulation data using principal component analysis (PCA) \cite{noor1980reduced}. Due to the limited expressiveness of linear subspaces, however, recent methods have resorted to neural networks that can directly capture the nonlinearity of the underlying physics.

\textbf{Neural Physics.}
Inspired by physics-informed neural networks (PINN) \cite{raissi2019physics}, current state-of-the-art methods \cite{d2022n, cloth, tretschk2020demea} leverage neural networks for subspace construction and simulation. These methods collect a large number of deformed states and train, e.g., deep autoencoders to obtain a latent manifold. A notable example is latent-space dynamics \cite{Fulton:LSD:2018}, which achieves physical awareness by collecting training data from a physical simulator. Shen et al. \cite{high-order-diff} augment latent-space dynamics with higher-order derivatives to accelerate simulation. Romero et al. \cite{10.1145/3450626.3459875} propose to learn nonlinear neural corrections for linear handle-based subspaces.
While these methods have shown the promise of learning-based subspaces, they rely on large amounts of curated simulation data for training. This data collection is time-consuming, and learned subspaces are biased towards the training data. %
Additionally, existing methods train networks in a purely geometric way, leading to subspaces that are riddled with high-energy configurations. By contrast, our method learns nonlinear subspaces in a physically principled and fully self-supervised way, eliminating the need for curated simulation data. Our learned subspaces are interpretable and exhibit superior regularity compared to previous methods. 
Sharp et al. \cite{sharp2023datafree} have likewise explored the potential of unsupervised learning for subspace generation. Their method introduces a regularization term to encourage mode orthogonality. However, as we show in our analysis (Sec. \ref{ssec:efficiency}), this regularizer cannot always prevent mode collapse, i.e., a decrease in effective dimensionality of the learned subspaces.

Neural simulation has attracted special attention for cloth simulation \cite{cloth, d2022n, 10.1145/3208159.3208162, 10.1145/3355089.3356512, 10.1145/3623264.3624441}. Bertiche et al. \cite{10.1145/3550454.3555491}, Santesteban et al. \cite{santesteban2022snug}, and Grigorev et al. \cite{grigorev2022hood} have investigated self-supervised learning techniques. Whereas these techniques are specialized for garments, our approach applies to a wider range of mechanical systems and materials.

\section{Theory}
We start by introducing our novel self-supervised learning approach for nonlinear physics-based subspaces in Section~\ref{ssec:selfsupervised}. 
Since our focus is on learning nonlinear modal subspaces, we briefly summarize linear modes and nonlinear extensions in Section~\ref{ssec:modal_subspaces}. In Section~\ref{ssec:neural_modes}, we present Neural Modes, i.e., learned nonlinear subspaces that are trained in a fully self-supervised way. 

\subsection{Self-Supervised Subspace Learning}
\label{ssec:selfsupervised}
We consider discrete elastic surfaces and solids represented as triangle meshes and tetrahedron meshes, respectively.
Let $\bx\in\mathcal{R}^{3n}$ and $\bX\in\mathcal{R}^{3n}$ denote vectors of deformed and undeformed positions for the $n$ vertices, respectively.
Equilibrium states of these discrete systems can be characterized by constrained energy minimization principles in the form of 
\begin{align}
    {\bf x}^*(\phi,\psi)=\textrm{arg}\min_{{\bf x}} \quad & E_{\phi}({\bf x}) \label{eqn:opt_ex}\\
    \textrm{s.t.} \quad & C_{\psi}({\bf x})={\bf 0} \nonumber
\end{align}
where $E_{\phi}({\bf x})$ is the mechanical energy defined through intrinsic parameters $\phi$ including material properties and input shape. Furthermore, $C_{\psi}({\bf x})$ is a vector-valued constraint function modeling the boundary conditions, which are defined through a set of extrinsic parameters ${\psi}$. 
The solution set ${\bf x}^*(\phi,\psi)$ constitutes a nonlinear subspace of $\mathcal{R}^{3n}$ corresponding to physically-principled equilibrium configurations. Using conventional simulation, sampling from this manifold amounts to solving a constrained minimization problem for each sample $(\phi_i,\psi_i)$, which is computationally expensive. Our approach to this problem is to find a neural network ${\bf x}[{\theta}^*](\phi,\psi)$ that best emulates the equilibrium subspace,
\begin{align}
    {\bf x}[{\theta}^*](\phi,\psi)\simeq\textrm{arg}\min_{\theta} \quad & E_{\phi}({\bf x}[{\theta}](\phi,\psi)) \label{eqn:opt_theta}\\
    \textrm{s.t.} \quad & C_{\psi}({\bf x}[{\theta}](\phi,\psi))={\bf 0} \ .\nonumber
\end{align}
Instead of optimizing for the unknown equilibrium configuration, a single neural network forward pass instantly provides the solution ${\bf x}^*$. Unlike supervised approaches that aim to learn solution spaces based on simulation data, we find the network weights $\theta$ by directly minimizing the mechanical energy across parameter space. To this end, we define a physics-based loss function,
\begin{align}
    L(\theta)= \mathbb{E}_{\phi,\psi}[\enspace E_{\phi}({\bf x}[{\theta}](\phi,\psi))+\lambda \|C_{\psi}({\bf x}[{\theta}](\phi,\psi))\|^2 \enspace] \ ,
    \label{eqn:loss_theta}
\end{align}
where we used a penalty function with stiffness coefficient $\lambda$ to enforce constraints. Once trained, the optimal $\theta$ give rise to a nonlinear subspace that, instead of approximating simulation data in a geometric sense, reflects the physical principles of the underlying mechanical system. Moreover, this learning process is entirely self-supervised and therefore easy to automate. We focus on thin shells and volumetric solids as discrete mechanical systems and build our subspaces on nonlinear compliant modes \cite{Duenser22NCM}. However, our formulation generalizes to a broad range of mechanical models, materials, and subspace parameterizations. The only requirement is that the energy function and constraints are differentiable and weakly convex. 
In the following, we apply our formulation to nonlinear modal subspaces.

\subsection{Modal Subspaces}\label{ssec:modal_subspaces}
\textbf{Linear Modes. } 
Equilibrium configurations of mechanical systems are minimizers of $E(\bx)$. To construct subspaces that contain physically plausible configurations, we seek directions that minimize the energy for given displacement magnitudes $z$, i.e.,
\begin{equation}
    \min_{\bu} E(\bX+ \bu) \quad \text{s.t.} \quad \tfrac12(\bu^T\bu-z)=0 \ .
\end{equation}
For small displacements, we can replace $E(\bX+\bu)$ with a quadratic approximation,
\begin{equation}
\label{eq:linearModesOptProblem}
    \min_{\bu} \tfrac12  \bu^T\bH\bu \quad \text{s.t.} \quad \tfrac12(\bu^T\bu-z)=0 \ ,
\end{equation}
where $\bH$ denotes the Hessian of the elastic energy and we have used the fact that both the energy and its gradient vanish at the origin. 
Considering the Lagrangian of this problem,
\begin{equation}
    \mathcal{L}=\tfrac12 \bu^T\bH\bu- \tfrac\mu2 (\bu^T\bu-z) \ ,
\end{equation}
where $\mu$ is a Lagrange multiplier, it becomes evident from the first-order optimality conditions,
\begin{align*}
    \bH\bu -\mu\bu \ &= {\bf 0} \ ,\\
    \bu^T\bu -z & = 0 \ ,    
\end{align*}
that solutions to the constrained optimization problem (\ref{eq:linearModesOptProblem}) are eigenvectors $\be_i$ of the Hessian, i.e., $\bH\be_i=\mu_i\be_i$.
In particular, each eigenvector $\be_{i+1}$ minimizes the system's energy in the $3n-i$-dimensional subspace orthogonal to all previous eigenvectors $\be_i$. This observation motivates the choice of constructing linear subspaces from the $m$ eigenvectors---the linear modes---corresponding to the smallest eigenvalues. 
For small deformations, where the quadratic energy approximation is accurate, linear modes form an energetically optimal subspace in the sense of the mechanical work required to achieve a displacement of given magnitude. For larger deformations, however, linear modes suffer from distortion artifacts that manifest as a steep increase in energy and physically-implausible configurations; see, e.g., Figure \ref{fig:armadillo}.

\textbf{Nonlinear Compliant Modes. }\label{ssec:nonlinear_modes}
To overcome the limitations of linear modes, D{\"u}nser et al. \cite{Duenser22NCM} define nonlinear compliant modes as
\begin{align}
\label{eq:NCM}
    {\bf n}_i(z) &= \textrm{arg}\min_{\bf u}E(\bX+{\bf u}) && \textrm{s.t.}\quad {\bf e}_i^T {\bf u}=z \ .
\end{align}
Letting ${\bf l}_i=z{\bf e}_i$ denote the displacement along the linear mode, we can alternatively rewrite Equation (\ref{eq:NCM}) as
\begin{align}
    {\bf n}_i(z) &= {\bf l}_i + \textrm{arg}\min_{\bf y}E(\bX+{\bf l}_i+{\bf y}) && \textrm{s.t.}\quad {\bf l}_i^T {\bf y}=0 \label{eqn:nonlinear_corr}
\end{align}
where the correction ${\bf y}$ is obtained by minimizing the system's energy in the space orthogonal to the linear mode. 
As demonstrated by D{\"u}nser et al., nonlinear compliant modes can achieve large and energetically-optimal deformations in the sense of (\ref{eq:NCM}). Nevertheless, the original formulation is restricted to individual modes and does not directly extend to multi-dimensional modal spaces. Furthermore, nonlinear compliant modes are very expensive to compute as evaluating the trajectory at a given parameter location requires the solution of a nonlinear constraint optimization problem. While these limitations seem to rule out nonlinear compliant modes for interactive subspace simulation, we show in the following that our self-supervised learning approach proposed in Section \ref{ssec:selfsupervised} can solve both problems in an elegant and efficient way.

\subsection{Neural Modes}\label{ssec:neural_modes}
Our goal is to build learned subspaces on the basis of nonlinear compliant modes. To this end, we first extend the concept by D{\"u}nser et al. to $m$-dimensional modal spaces,
\begin{align}
\label{eqn:opt_neural_full}
    {\bf n}(\bz) &= {\bf l} + \textrm{arg}\min_{\bf y}E(\bX+{\bf l}+{\bf y}) && \textrm{s.t.}\quad {\bf l}^T {\bf y}=0 \ , 
\end{align}
where the linear displacement ${\bf l}=\sum_i^m z_i {\bf e}_i$ is now a combination of linear modes weighted by the modal coordinates $\bz=(z_1,\ldots,z_m)$. 
We then train a neural network to learn the solution to the above optimization problem for \textit{any} choice of modal coordinates, i.e.,
\begin{align}
    {\bf n}[\theta^*](\bz)={\bf l}+\by[\theta^*](\bz)\\
    \theta^*=\textrm{arg}\min_{{\theta}} \ E(\bX+{\bf l}+{\bf y}[{\theta}](\bz)) \\
    \nonumber
    \textrm{s.t.} \enspace {\bf l}^T {\bf y}[\theta](\bz) =0 \ .
    \end{align}
We refer to the resulting nonlinear modal subspace $\bn[\theta](\bz)$ as \textit{Neural Modes}.
To train Neural Modes, we use the above expression in 
\eqref{eqn:loss_theta} and obtain the loss function for self-supervised learning as
\begin{align}
\label{eqn:loss_neural}
    L(\theta) =\mathbb{E}_{\bz}[\enspace E(\bX+{\bf l}+{\bf y}[{\theta}](\bz)) 
     + \lambda ({\bf l}^T {\bf y}[{\theta}](\bz))^2 \enspace ]
\end{align}
During training, we draw the modal coordinates $\bz$ from a uniform distribution with predefined domain. We input $\bz$ to the network, and retrieve vertex positions $\bx=\bX+{\bf l}+{\bf y}$ from the network output $\bf y$. While experimenting with this loss function, we observed that the trained network produces nonzero corrections at the origin, i.e., $\by[\theta](\bzero)\neq\bzero$. Since Neural Modes should coincide with linear modes for small displacements, we add a term penalizing corrections at the origin, leading to the modified loss function
\begin{align}
    \label{eqn:loss_final}
    L_{\textrm{final}}(\theta)=L(\theta)+\eta \|{\bf y}[{\theta}]({\bf 0})\|^2 \ .
\end{align}

\textbf{Hyperparameters. }
The formulation described so far computes Neural Modes for given material and shape parameters. Nevertheless, these parameters significantly affect the resulting nonlinear modes. Fortunately, we can directly learn the effect of these changes by simply adding corresponding parameters to modal coordinates and sampling during training.

\subsection{Subspace Dynamics}\label{ssec:dynamics}
The goal of subspace simulation is to create compelling motion with physically plausible deformations at a fraction of the cost associated with full-space simulation. Starting from a variational re-formulation of the implicit Euler update equations in full space \cite{example2}, we have
\begin{align}
 \label{eqn:dyn_u}
    {\bf u}_{n+1} = \textrm{arg}\min_{\bf u} \frac{1}{2h^2}\| {\bf u}-2{\bf u}_n+{\bf u}_{n-1} \|_\bM^2+E({\bf u}) \ ,
\end{align}
where $h$ is the step size and $||\cdot||_\bM$ is the metric induced by the mass matrix $\bM$. Following the formulation of Fulton et al.  \cite{Fulton:LSD:2018}, we replace the full-space displacement ${\bf u}$ with our neural subspace ${\bf n}({\bf z})$ as
\begin{align}
    \label{eqn:dyn_z}
    {\bf z}_{n+1} = \textrm{arg}\min_{\bf z} \frac{1}{2h^2}\| {\bf n}({\bf z})-2{\bf u}_n+{\bf u}_{n-1} \|_M^2+E({\bf n}({\bf z})) \ ,
\end{align}
We solve the minimization problem for end-of-step modal coordinates using L-BFGS \cite{liu1989limited}.

\section{Results}
\label{sec:results}

We implemented all our experiments and network training in PyTorch, taking advantage of automatic differentiation. We use conventional multi-layer perceptrons (MLP) with Gaussian Error Linear Units (GELU) to implement our Neural Modes. 
We also experimented with Residual Networks (ResNet) but found no improvements in practice. 
We train our networks on a single RTX 3090 GPU using L-BFGS with line search as the optimizer. We use discrete shells \cite{grinspun2003discrete} for deformable surfaces and linear tetrahedron elements with a Saint Venant-Kirchhoff material for solids \cite{Kim22FEMCourse}. We summarize our experiment setup in Table~\ref{tab:setup}.
\begin{table*}
\caption{Overview of our example setups. \label{tab:setup}}
\vspace{-0.2cm}
\begin{minipage}{\linewidth}
\begin{center}
\resizebox{0.99\columnwidth}{!}{
\begin{tabular}{lcccccccccc}
  \toprule
  Model & Energy & \# of cells & Full dim & Reduced dim & Neural dim & Auxiliary dim & MLP & $\lambda$ & $\eta$ & Timestep\\
  \midrule
  Sheet & Discrete Shell & 100 & 363 & 3 & 3 & 0 & 5x64 & 1e8 & 1e7 & 40 ms \\
  Sheet (aspect ratio) & Discrete Shell & 100 & 363 & 4 & 4 & 1 & 8x64 & 1e8 & 1e7 & -- \\
  Giraffe & Tetrahedron+Penalty & 5424 & 5094 & 3 & 3 & 0 & 10x64 & 1e8 & 1e7 & 36 ms\\ 
  Armadillo & Tetrahedron & 26211 & 21537 & 11 & 5 & 0 & 7x256 & 1e8 & 1e7 & 35 ms\\
  Bunny & Tetrahedron+Penalty & 7585 & 6360 & 8 & 8 & 0 & 8x64 & 1e6 & 1e3 & 28 ms\\
  \bottomrule 
\end{tabular}
}
\end{center}
\end{minipage}

\vspace{-0.2cm}
\end{table*}

\subsection{Performance Comparisons}
We quantitatively evaluate our approach on a square shell with a flat rest state. We train a 5-layer MLP to learn a 3-dimensional modal subspace using modal coordinates ${\bf z}$ from the domain $[-0.625, 0.625]^3$. We compare two variants of training for our MLP using stochastic and regular (grid-based) sampling, respectively. For stochastic sampling, we choose different modal coordinates ${\bf z}$ in each iteration by uniformly sampling the domain $[-0.625, 0.625]^3$. For regular sampling, we sample ${\bf z}$ from a $9^3$ uniform grid. 

We compare our method with Latent-space Dynamics \cite{Fulton:LSD:2018}, which combines a deep autoencoder (AE) with principal component analysis (PCA). We abbreviate their method as PCA+AE. Our MLP essentially corresponds to the decoder part of the autoencoder used in PCA+AE. For a fair comparison, we use the same MLP as PCA+AE's decoder and make a symmetric copy for the encoder.

 To train PCA+AE, we collect a dataset by regularly sampling ${\bf z}$ and solving (\ref{eqn:opt_neural_full}) for each sample. The training, validation, and test sets have resolutions of $9^3$, $7^3$, and $11^3$ samples, respectively.
We also collect a second training set of the same size by randomly sampling modal coordinates. We train separate instances of PCA+AE on these two training sets (grid, random), and train our MLP using both stochastic and regular sampling. We evaluate and compare all methods on the same test and validation sets.

\begin{table*}
\caption{Performance comparison on the square sheet example. We compute the elastic energy $E$ and its error $\Delta E$ with respect to ground truth on both validation and test sets. The subscript denotes the reduction operator for the batch dimension.\label{tab:energy}}
\vspace{-0.15cm}
\centering
\begin{minipage}{0.8\linewidth}
\begin{center}
\resizebox{\columnwidth}{!}{
\begin{tabular}{@{\extracolsep{3pt}} lcccccccccccc}
  \toprule
  \multirow{2}{*}{Method} & \multicolumn{4}{c}{Validation} & \multicolumn{4}{c}{Test}\\
  \cline{2-5} \cline{6-9}
      & $\Delta E_{avg}$ & $\Delta E_{max}$ & $\Delta E_{std}$ & $E_{avg}$ & $\Delta E_{avg}$ & $\Delta E_{max}$ & $\Delta E_{std}$ & $E_{avg}$\\ 
  \midrule
  PCA+AE           & 1814 & 22225 & 2005 & 18144 & 2259 & 27991 & 2646 & 24007 \\
  PCA+AE (grid)    & 2188 & 36168 & 3626 & 18088 & 2649 & 72834 & 4664 & 23217  \\
  L2 Supervised    & 1630 & 10231 & 1148 & 18210 & 2074 & 13750 & 1697 & 24325 \\
  Neural Modes (stoch.)   & \bf 376 & \bf 1197 & \bf 170 & \bf 17032 & \bf 386 & \bf 1154 & \bf 183 & \bf 22891 \\
  Neural Modes (grid)       & \bf 199 & \bf 754 & \bf 118 & \bf 16831 & \bf 245 &\bf 3062 & \bf 215 & \bf 22730 \\
  \bottomrule
\end{tabular}
}
\end{center}
\end{minipage}

\vspace{-0.3cm}
\end{table*}

\begin{table}
\caption{We compare early stopping criteria using elastic energy and geometry-based L2 loss as metrics. \label{tab:early_stop}}
\vspace{-0.1cm}
\begin{minipage}{\columnwidth}
\begin{center}
\resizebox{0.9\columnwidth}{!}{
\begin{tabular}{@{\extracolsep{3pt}} lcccc}
  \toprule
  \multirow{2}{*}{Method} & \multicolumn{2}{c}{Validation} & \multicolumn{2}{c}{Test}\\
  \cline{2-3} \cline{4-5}
     & L2 & Energy & L2 & Energy\\ 
  \midrule
  PCA+AE           & 18144 &\bf 17762& 24007 &\bf 23484 \\
  PCA+AE (grid)    & 18088 &\bf 18042 & 23217 &\bf 23051\\
  L2 Supervised    & 18210 &\bf 17690 & 24325 &\bf 23682 \\
  \bottomrule
\end{tabular}
}
\end{center}
\end{minipage}

\vspace{-0.2cm}
\end{table}

\begin{table*}
\caption{Comparison of internal stress on the square sheet example. We compare average per-element stresses weighted by areas and maximum per-element stress. We compute the Frobenius norm of stress $|S|$ and its error $\Delta |S|$ with respect to ground truth values for each method. The subscript denotes the reduction operator for the batch dimension.\label{tab:stress}}

\vspace{-0.15cm}
\centering
\begin{minipage}{0.8\linewidth}
\begin{center}
\resizebox{\columnwidth}{!}{
\begin{tabular}{@{\extracolsep{3pt}} lcccccccccccc}
  \toprule
  \multirow{2}{*}{Method ($\times 10^6$)} & \multicolumn{4}{c}{Weighted by elements} & \multicolumn{4}{c}{Max over elements}\\
  \cline{2-5} \cline{6-9}
      & $\Delta |S|_{avg}$ & $\Delta |S|_{max}$ & $\Delta |S|_{std}$ & $|S|_{avg}$ & $\Delta |S|_{avg}$ & $\Delta |S|_{max}$ & $\Delta |S|_{std}$ & $|S|_{avg}$\\
  \midrule
  PCA+AE            & 25 & 138 & 17 & 251 & 67 & 431 & 49 & 520 \\
  PCA+AE (grid)    & 25  & 305 & 29 & 245 & 64 & 784 & 68 & 505 \\
  L2 Supervised    & 25 & 100 & 14 & 255 & 87 & 350 & 57 & 550  \\
  Neural Modes (stoch.)   & \bf 9 & \bf 36 & \bf 7 & \bf 240 & \bf 22 & \bf 155 & \bf 21 & \bf 486 \\
  Neural Modes (grid)       & \bf 6 & \bf 24 & \bf 5 & \bf 236 & \bf 16 & \bf 124 & \bf 16 & \bf 479 \\
  \bottomrule
\end{tabular}
}
\end{center}
\end{minipage}
\vspace{-0.2cm}
\end{table*}

\textbf{Elastic Energy. }
The evaluation results are shown in Table~\ref{tab:energy}. Since nonlinear modes are constrained minimizers of (\ref{eqn:opt_neural_full}), we use average elastic energy as a metric for measuring the accuracy of a subspace.
It can be seen from Table~\ref{tab:energy} that our method has lower average energy than PCA+AE and an order of magnitude smaller errors compared to the ground truth solution. We hypothesize that the lower accuracy of PCA+AE is due to its geometry-based loss function and the associated likelihood of overfitting to the training set. 
We test this hypothesis in two ways.

\begin{figure}[h]
\begin{minipage}{\columnwidth}
    \centering
    \includegraphics[width=\linewidth]{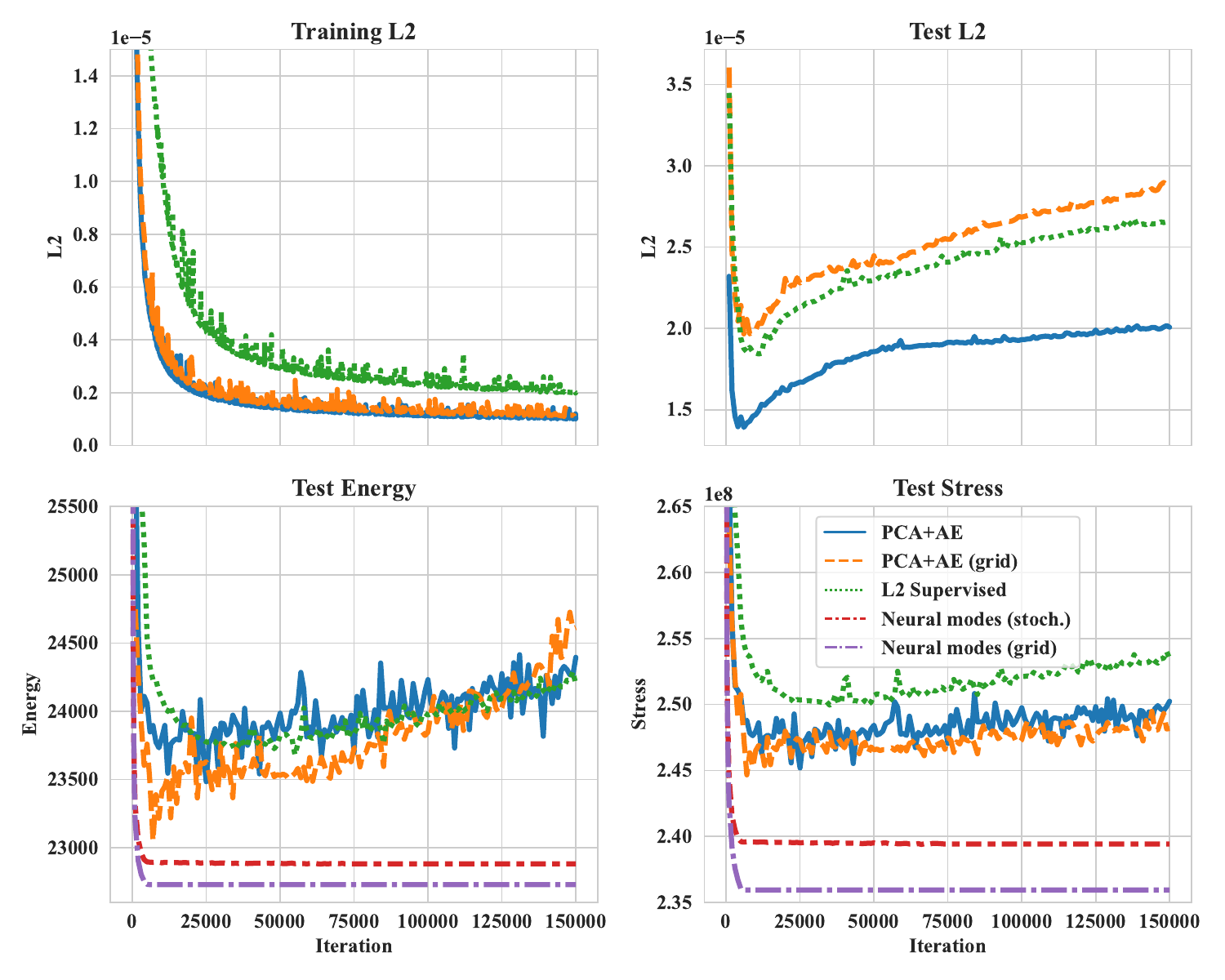}
    \vspace{-0.6cm}
    \caption{Supervised learning leads to overfitting for all loss functions, i.e., geometry-based L2 distance, average energy $E_{avg}$, and average stress $|S|_{avg}$. In contrast, our Neural Modes do not suffer from overfitting.}
    \label{fig:plot_metrics}
\end{minipage}
\vspace{-0.5cm}
\end{figure}

First, we train our network again using the same architecture but use the loss function from PCA+AE, which measures the geometric distance to the training data in the L2 sense. We then train on the same regular sampling set, i.e., using the same data as before. As can be seen from Table \ref{tab:energy}, the resulting network (\textit{L2 supervised}) exhibits significantly higher energy than our Neural Modes.
Second, we plot learning curves for all networks in Figure~\ref{fig:plot_metrics}. We can see that, while PCA+AE and L2 Supervised variants reduce training L2 loss steadily, their test L2 loss, energy, and stress all increase during training. In contrast, our self-supervised Neural Modes do not suffer from overfitting, steadily reducing the energy and stress of the test set.
Finally, we experimented with early stopping, which is a common strategy to mitigate overfitting. We use both L2 loss and elastic energy as stopping criteria and observed somewhat better results for the PCA+AE variants when using energy (see Table~\ref{tab:early_stop}). Nevertheless, the energy for PCA+AE is still significantly higher than for Neural Modes. 

\textbf{Stress \& Nodal Force. }
We further evaluate the learned subspace by analyzing internal stresses and nodal forces. We compute averaged stress weighted over element areas and maximum element stress in Table~\ref{tab:stress}. We achieve smaller stress and significantly lower errors compared with supervised methods. In addition to stress, we also observe one order of magnitude smaller errors in nodal forces (see Table~\ref{tab:nodal_force}).
These results support our hypothesis that the geometric L2 loss is unable to capture the physical principles that govern the manifold of equilibrium configurations. It is furthermore prone to overfitting and can be expected to generalize poorly in practice. By directly minimizing energy during training, our self-supervised method shows much better performance while avoiding overfitting problems.

\textbf{Deformable Solids. }
We additionally evaluate the quality of Neural Modes for deformable solids, using the example shown in Figure~\ref{fig:armadillo}. We stochastically train an MLP to learn a 5-dimensional subspace. Similar to the previous example, we collect 1300 equilibrium states by uniformly sampling modal space and compute ground truth solutions for each sample. The first 900 samples are used for training, the subsequent 100 samples are used for validation, and the last 300 samples are reserved for the test set. 
We stop training after 2050 epochs as the L2 loss evaluated on the validation set continually increases for the following 500 epochs. The statistics are shown in Table~\ref{tab:armadillo_energy_int}. 
To ensure our evaluation result is not sampled by chance, we break down the test set into subintervals and compare Neural Modes and PCA+AE on each subinterval in addition to the complete set. It can be seen that Neural Modes outperform PCA+AE both for every interval and for the complete test set. This result confirms our observation for deformable surfaces, indicating that Neural Modes consistently outperform PCA+AE by significant margins. 

\begin{table}
\caption{We evaluate the performance of our method using the model's average elastic energy evaluated on the test set.}
\vspace{-0.2cm}
\label{tab:armadillo_energy_int}
\setlength\tabcolsep{2.3pt}
\begin{minipage}{\columnwidth}
\begin{center}
\resizebox{\columnwidth}{!}{
\begin{tabular}{lcccccc}
  \toprule
  Method ($\times 10^4$) & 0$\sim$20\% & 20$\sim$40\% & 40$\sim$60\% & 60$\sim$80\% & 80$\sim$100\% & all \\
  \midrule
  Ground truth     & 12.3 & 9.7 & 12.5 & 10.9 & 10.8 & 11.3\\
  PCA+AE  & 106.7 & 4329.9 & 216.5 & 179.5 & 104.8 & 987.5\\ 
  Neural Modes  & \bf 57.1 & \bf 58.0 & \bf 58.9 & \bf 52.3 & \bf 51.9 & \bf 55.8\\
  \bottomrule
\end{tabular}
}
\end{center}
\end{minipage}

\vspace{-0.4cm}
\end{table}

\subsection{Latent Space Structure}
\begin{figure*}

\begin{minipage}{\linewidth}
\begin{center}
\resizebox{\columnwidth}{!}{

\includegraphics[width=\linewidth]{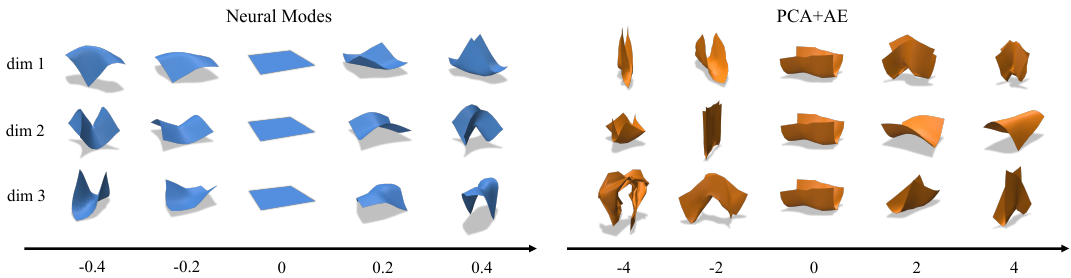}

}
\end{center}
\end{minipage}
\vspace{-0.2cm}
\caption{Visualization of latent space dimensions for Neural Modes and PCA+AE.\label{fig:fabric_vis}}
\vspace{-0.4cm}
\end{figure*}

The previous experiments indicate that our self-supervised learning approach yields subspaces with lower mechanical energy, stress, and nodal force than PCA+AE. Consequently, Neural Modes exhibit substantially lower errors with respect to ground truth values obtained by solving the underlying constrained minimization problem.
We further analyze the structure of the latent spaces with a focus on smoothness, interpretability, and regularity. Specifically, we investigate whether the map from latent space to full space is (1) \textit{smooth} in the sense that smooth motion in latent space leads to smooth interpolated trajectories in geometry space, (2) \textit{regular} in the sense that interpolated geometry is semantically bounded by terminal states, and (3) \textit{interpretable} in the sense that individual latent space dimensions have a clear and distinguishable effect on the full space geometry.

To analyze interpretability, we consider the same example as before and explore the behavior of Neural Modes and PCA+AE when walking along individual latent space coordinate directions. As can be seen from Fig. \ref{fig:fabric_vis}, Neural Modes faithfully reproduce the individual nonlinear modes and preserve symmetry. We also observe physically-plausible deformations when combining multiple latent space dimensions (see Figure \ref{fig:arithmetic_vis}). The picture looks quite different for PCA+AE. Although each latent space dimension features meaningful shapes, they are interspersed with distorted or crumpled shapes that are physically implausible. We further note that Neural Modes preserve both the origin (i.e., $\bn(\bzero)=\bX$) and symmetry of the ground truth modes, whereas PCA+AE does neither.

A basic test for regularity is interpolation quality \cite{ali2021evaluation}. The median of two latent vectors is expected to yield a semantically meaningful median result and no abrupt changes from terminal states. We visualize interpolation quality in Figures~\ref{fig:fabric_vis} and \ref{fig:arithmetic_vis} using evenly spaced sample points. It can be seen that PCA+AE exhibits poor regularity. By contrast, Neural Modes generate semantically meaningful intermediate states, defining a smooth trajectory in geometry space. Unlike PCA+AE, this semantic interpolation property makes Neural Modes attractive for applications such as keyframe animations (see Figure~\ref{fig:keyframe}).

From these experiments, we conclude that, compared to supervised learning from simulation data (PCA+AE), self-supervised learning of Neural Modes yields well-behaved subspaces with much better smoothness, interpretability, regularity, and structure preservation.

\subsection{Material and Shape Parameters}
Neural modes are naturally parameterized by modal coordinates, but they generalize to other parameters such as shape and material properties. Figure~\ref{fig:latent_shape} showcases a simple example where we add aspect ratio as an auxiliary parameter for a rectangular sheet. Starting from a sheet with edge length ratio 2:1, the folding direction smoothly changes from the sheet's short axis to its diagonal as we decrease the ratio towards 1:1. 

\subsection{Subspace Efficiency} \label{ssec:efficiency}
\begin{figure}[t]
\vspace{0.2cm}
\begin{minipage}{\columnwidth}
    \centering
    \includegraphics[width=0.8\columnwidth]{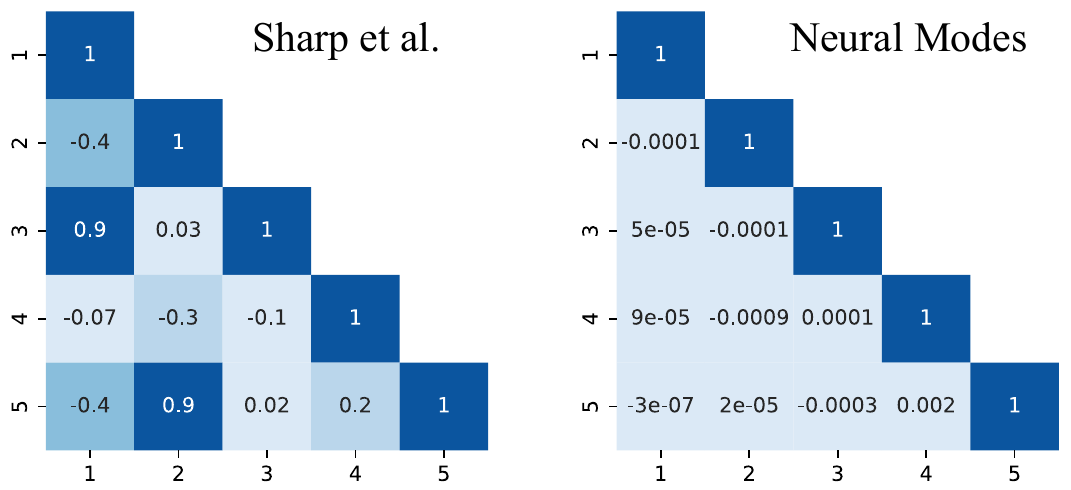}
    \vspace{-0.2cm}
    \caption{Correlation matrices of latent directions for Armadillo.}
    \label{fig:corr}
\end{minipage}
\vspace{-0.8cm}
\end{figure}

We additionally compare subspace efficiency with Sharp et al. \cite{sharp2023datafree}. Like our method, Sharp et al. learn nonlinear subspaces in an unsupervised way using the mechanical energy $E({\bf x})$ as loss function.  Using only energy, however, the optimal network would map all latent variables to the undeformed state $\bx=\bX$ since $E({\bf X})=0$. To avoid this \textit{subspace collapse}, Sharp et al. introduce a regularization term that encourages the network to produce an isometric map from latent space to full space. Although this regularizer avoids subspace collapse, it cannot prevent \textit{mode collapse}, i.e., a loss of effective dimensionality. 
To show this, we use the method by Sharp et al. to learn a 5D-subspace for the Armadillo model. We compute latent directions $\mathbf{E}=[\be_1|\ldots|\be_5]$ by evaluating the network's gradient at the origin and compute the correlation matrix $\bE^T\bE$ (Fig. \ref{fig:corr}). The eigenvalues of the correlation matrix are $(2.3, 1.6, 1.1, 0.0,0.0)$, revealing that the subspace is effectively 3-dimensional. Applying the same test to Neural Modes yields first-order orthogonal latent directions and a quasi-diagonal correlation matrix with all eigenvalues close to $1.0$ (see Figure \ref{fig:corr}). 

Mode collapse limits the efficiency of subspaces. As shown in Figure \ref{fig:armadillo_vis}, while our Neural Modes capture both arm and leg motions, the method by Sharp et al. only learns leg motions with the aforementioned redundancy.

\subsection{Applications}
Complementing our quantitative analysis, we additionally evaluate our method on a set of application examples that illustrate potential use cases in modeling and animation.

\begin{figure}
\vspace{-0.2cm}
\begin{minipage}{\columnwidth}
    \centering
    \includegraphics[width=\columnwidth]{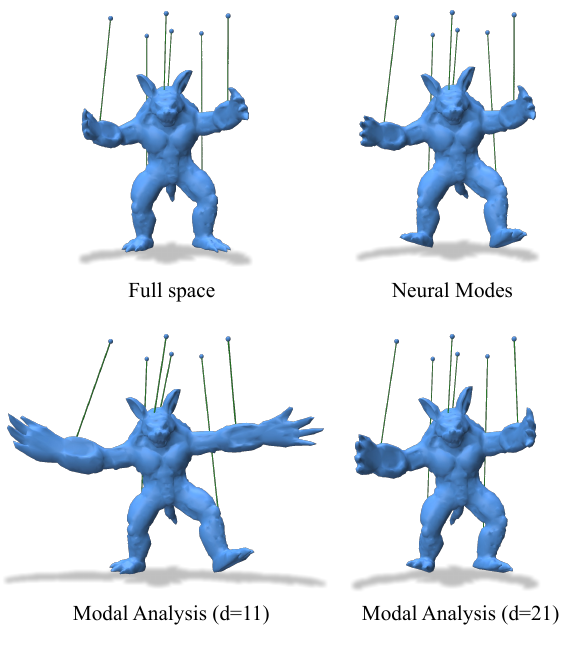}
    \vspace{-0.6cm}
    \caption{Comparison of 5d Neural Modes, modal analysis, and full-space simulation on an Armadillo puppet. 
    }
    \label{fig:armadillo}
\end{minipage}
\vspace{-0.3cm}
\end{figure}

\textbf{Subspace Dynamics}
As our first example, we consider a subspace simulation problem using the deformable Armadillo puppet shown in Figure~\ref{fig:armadillo}. The model is driven by six rods whose rest lengths we animate using sinusoidal motion. We train an 11-dimensional subspace, with 5 DoFs corresponding to Neural Modes, and 6 DoFs for rigid transformations.
It is evident from Figure~\ref{fig:armadillo}---but best observed in the accompanying video---that linear modes lead to large distortion artifacts in the arms and legs. This issue is only partly mitigated when adding ten extra basis vectors. Neural Modes, in contrast, produce nonlinear rotations as expected from the ground truth simulation. Moreover, our method achieves real-time performance in $\sim$35ms per time step, whereas the full simulation is two orders of magnitude slower. 

We consider two additional examples to illustrate that our method generalizes to arbitrary shapes: a giraffe and a bunny model simulated using Neural Modes subspaces with 3 and 8 dimensions, respectively.
Both models run at real-time rates and produce smooth, visually pleasing motion. Please see the accompanying video for animations.

\textbf{Keyframing. }
\begin{figure}
\begin{minipage}{\columnwidth}
    \centering
    \includegraphics[width=\linewidth]{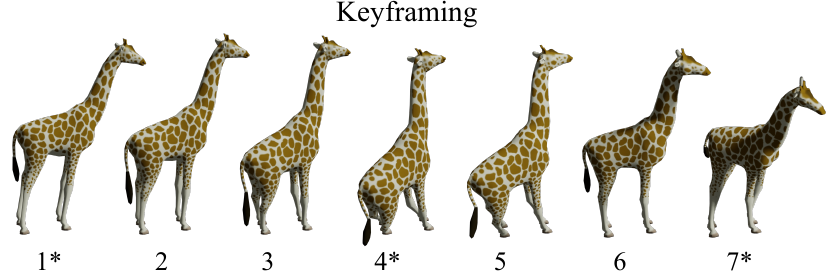}
    \vspace{-0.5cm}
    \caption{Keyframing of nonlinear sitting-down and standing-up motions using Neural Modes. Asterisks denote keyframes.}
    \label{fig:keyframe}
\end{minipage}
\vspace{-0.5cm}
\end{figure}
Keyframe animation is a basic tool for generating motion from a sparse set of input poses. Given a set of keyframe poses through modal coordinates, we can generate in-between trajectories with physically plausible deformations by means of interpolation. 
Figure~\ref{fig:keyframe} illustrates this application on the giraffe model.
While linear subspaces typically require nonlinear interpolation schemes to achieve acceptable results, our nonlinear subspaces produce physically plausible deformations even for simple linear interpolation of modal coordinates.

\section{Conclusion}

We presented a novel self-supervised method for learning nonlinear physics-based subspaces for real-time simulation. We demonstrated the potential of our formulation by learning Neural Modes, i.e., physics-based subspaces built on a nonlinear extension of linear eigenmodes. We quantitatively evaluated our method on a set of examples that include thin shells and volumetric solids. Furthermore, we performed an extensive ablation study, revealing that directly minimizing elastic energy leads to significantly better physical accuracy than using simpler geometry-based loss functions. We also showed that self-supervised learning resolves overfitting problems that supervised approaches often struggle with. By visualizing latent spaces, we demonstrated that Neural Modes are interpretable, free from singularities, and well-behaved for arbitrary modal combinations. We further showed that, unlike existing alternatives, our formulation avoids the problem of mode collapse. Finally, we showcase applications of our approach to subspace simulation and keyframing.

\subsection*{Limitations \& Future Work. }
Our learning method is restricted to differentiable energy functions and constraints.
Additionally, we did not investigate simulation acceleration algorithms and relied on L-BFGS for optimization. Future explorations could compute the exact Hessian in the reduced parameter space, which would open the door to more efficient minimization algorithms such as the Newton-Raphson method. 
While we have focused on subspaces induced by nonlinear modes, our approach generalizes to many other parameterizations. An exciting avenue for future work in this context would be to extend our method to inverse design problems, in, e.g., architecture or engineering design.

\newpage
{
    \small
    \bibliographystyle{ieeenat_fullname}
    \bibliography{main}
}

\clearpage
\setcounter{page}{1}
\appendix
\setcounter{table}{0}
\setcounter{figure}{0}
\renewcommand{\thetable}{A\arabic{table}}
\renewcommand\thefigure{A\arabic{figure}}
\renewcommand{\theHtable}{A.Tab.\arabic{table}}
\renewcommand{\theHfigure}{A.Abb.\arabic{figure}}
\renewcommand\theequation{A\arabic{equation}}
\renewcommand{\theHequation}{A.Abb.\arabic{equation}}

\newpage
\twocolumn[
        \centering
        \Large
        \textbf{\thetitle}\\
        \vspace{0.5em} Appendix \\
        \vspace{1.0em}
       ]

\section{Additional Tables and Figures}
\begin{table}[h]
\caption{Analysis of nodal forces on the square sheet example. We compute L1 and L2 norms of nodal forces $F$ and corresponding errors $\Delta F$ with respect to ground truth. The norms are normalized with respect to batch size. \label{tab:nodal_force}}
\vspace{-0.2cm}
\begin{minipage}{\columnwidth}
\begin{center}
\resizebox{\columnwidth}{!}{
\begin{tabular}{@{\extracolsep{3pt}} lcccccccccccc}
  \toprule
  {Method ($\times 10^4$)}
      & $\|\Delta F\|_{1}$ & $\|\Delta F\|_{2}$ & $\|F\|_{1}$ & $\| F\|_{2}$\\ 
  \midrule
  PCA+AE           & 111 & 5.95 & 279 & 24\\
  PCA+AE (grid)    & 83 & 4.35 & 250 & 22   \\
  L2 Supervised    & 125 & 6.59 & 293 & 24  \\
  Neural Modes (stoch.)   & \bf 31 & \bf 0.93 & \bf 199 & \bf 19   \\
  Neural Modes (grid)       & \bf 25 & \bf 0.74 & \bf 193 & \bf 19 \\
  \bottomrule
\end{tabular}
}
\end{center}
\end{minipage}

\vspace{-0.3cm}
\end{table}

\begin{figure}[h]

\begin{minipage}{\columnwidth}
    \centering
    \includegraphics[width=\columnwidth]{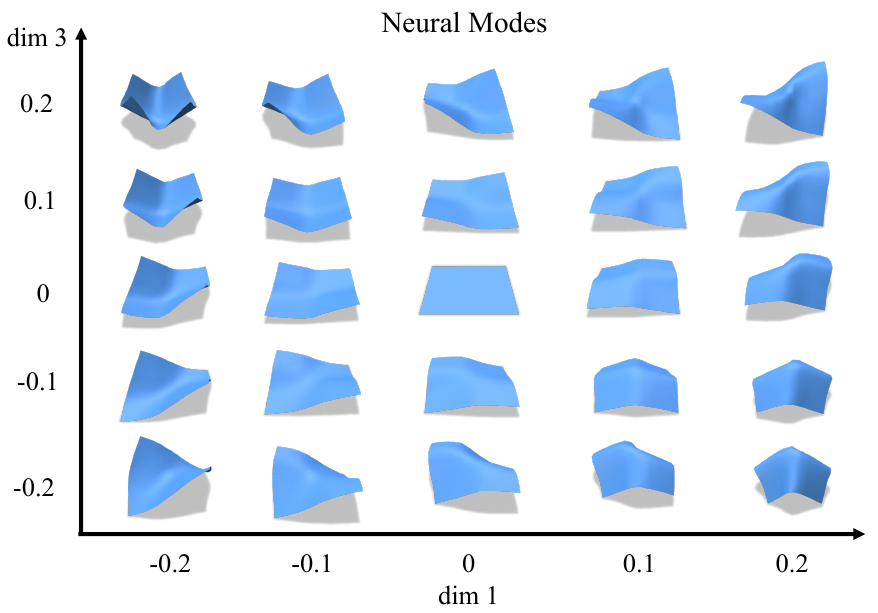}
    \vspace{-0.5cm}
    \caption{Visualization of a two-dimensional Neural Modes slice for the square sheet example.}
    \label{fig:arithmetic_vis}
\end{minipage}
\vspace{-0.2cm}
\end{figure}
\begin{figure}[h]
\vspace{0.1cm}
\begin{minipage}{\columnwidth}
    \centering
    \includegraphics[width=\columnwidth]{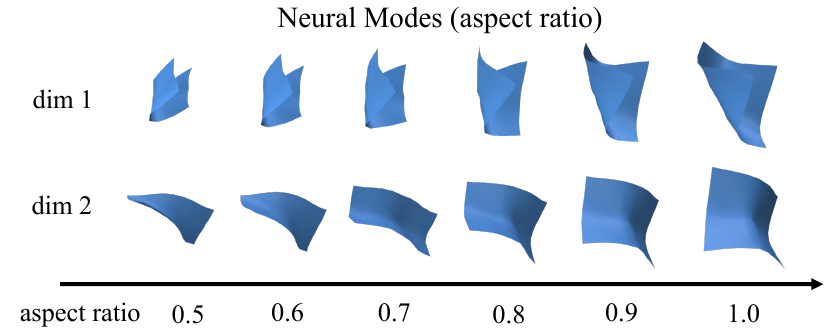}
    \caption{Visualization of Neural Modes for a rectangular sheet conditioned on different side length ratios.}
    \label{fig:latent_shape}
\end{minipage}
\vspace{-0.3cm}
\end{figure}

\begin{figure}[t]
\vspace{-15cm}
\begin{minipage}{\columnwidth}
    \centering
    \includegraphics[width=\columnwidth]{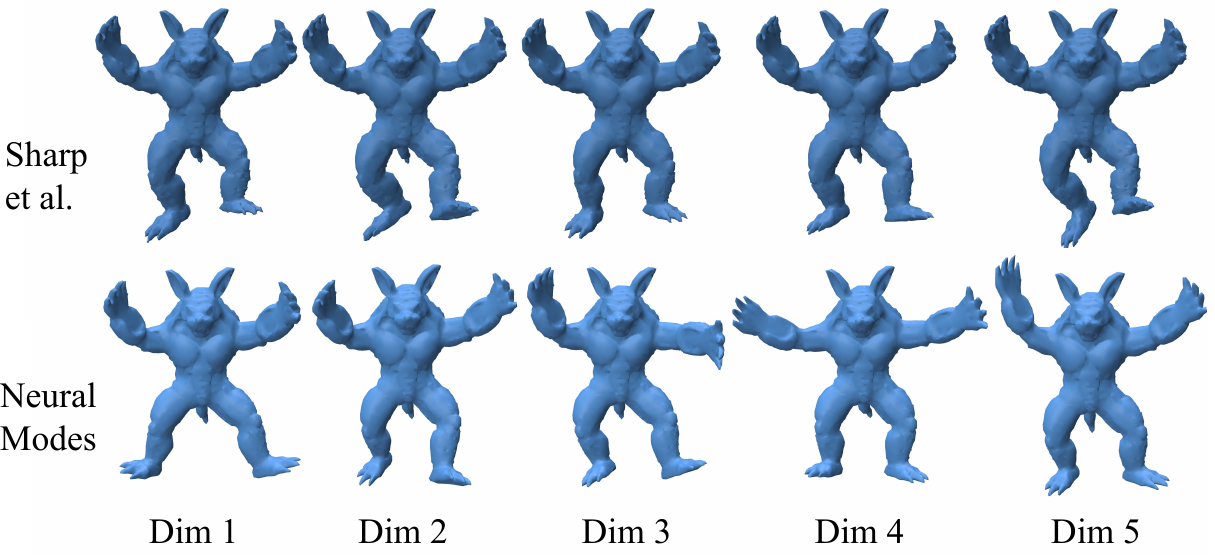}
    \vspace{-0.6cm}
    \caption{Visualization of Armadillo subspace ($d=5$)}
    \label{fig:armadillo_vis}
\end{minipage}
\vspace{-0.4cm}
\end{figure}

\end{document}